\documentclass[letterpaper, 10 pt, conference]{ieeeconf}
\IEEEoverridecommandlockouts 
\usepackage{graphicx} 
\usepackage{textcomp}
\usepackage{gensymb}
\usepackage{epsfig} 
\usepackage{xcolor}
\usepackage{mathptmx} 
\usepackage{times} 
\usepackage{amsmath} 
\usepackage{amssymb}  
\usepackage{acronym}
\usepackage{soul}
\usepackage{authblk}
\usepackage{subfloat}
\usepackage{subcaption}

\usepackage{booktabs}
\usepackage{array}{}
\usepackage{multirow}

\usepackage{isomath}

\usepackage{arydshln} 
\usepackage[linesnumbered,ruled,vlined]{algorithm2e}

\usepackage{ulem}
\usepackage{units}
\newcommand{\subparagraph}{}

\usepackage{todonotes}

\title{\LARGE \bf Shape, Size, and Fabrication Effects in 3D Printed Granular Jamming Grippers}

\author{David Howard$^{1}$, Jack O'Connor$^{1,2}$, James Brett$^{1}$, Gary W. Delaney$^{3}$%
\thanks{$^{1}$Robotics and Autonomous System Group, Data61, CSIRO, Australia {\tt\small david.howard@csiro.au}}%
\thanks{$^{2}$School of Mechanical and Mining, University of Queensland, Australia}%
\thanks{$^{3}$Computational Modelling Group, Data61, CSIRO, Australia }%

}

\begin{document}

\maketitle


\begin{abstract}
Granular jamming is a popular soft actuation mechanism that provides high stiffness variability with minimum volume variation.  Jamming is particularly interesting from a design perspective, as a myriad of design parameters can potentially be exploited to induce a diverse variety of useful behaviours.  To date, grain shape has been largely ignored.  Here, we focus on the use of 3D printing to expose design variables related to grain shape and size.  Grains are represented by parameterised superquadrics (superellipsoids); four diverse shapes are investigated along with three size variations.  Grains are 3D printed at high resolution and performance is assessed in experimental pull-off testing on a variety of benchmark test objects.  We show that grain shape and size are key determinants in granular gripping performance.  Moreover, there is no universally-optimal grain shape for gripping.  Optical imaging assesses the accuracy of printed shapes compared to their ideal models. Results suggest that optimisation of grain shape is a key enabler for high-performance, bespoke, actuation behaviour and can be exploited to expand the range and performance of granular grippers across a range of diverse usage scenarios.

\end{abstract}

 \section{INTRODUCTION}
\label{sec:introduction}

\begin{figure}[h!]
\centering
\includegraphics[width=0.9\columnwidth]{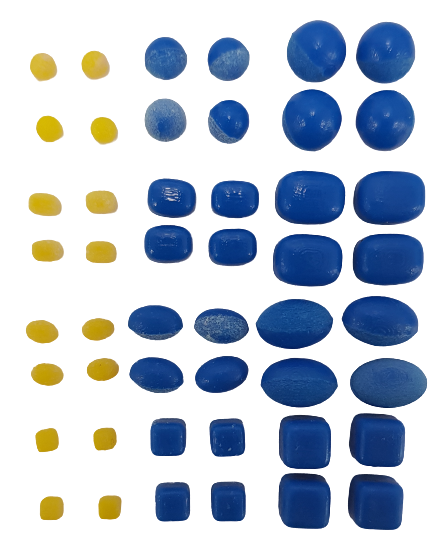}
\caption{The printed grains used in the study, in groups of 4 per combination of size and shape. Left-Right: 3mm Sphere-Volume Equivalent (SVE), 5SVE, 7SVE. Top-bottom: sphere, superellipsoid, ellipsoid, superball.}
\label{fig:fig1}
\end{figure} 

Design of soft actuators is a topic of increasing interest within the research community, as intrinsic properties of the gripper, i.e., compliance, stiffness variation, can be fully harnessed to maximise task performance~\cite{hughes2016soft,shintake2018soft}.  At a high level, this process involves iteration through a selection of candidate materials and geometries to arrive within a suitable performance envelope, which can involve elements of modelling as well as physical iterate-and-test, depending on the type of gripper under consideration.

Granular jamming actuators are a particularly promising soft actuator as they are comparatively highly parameterised thanks to the potential consideration of tunable grain properties (shape, size, material), alongside more ubiquitous membrane (material and geometry) and control properties~\cite{jamming_review}.  The inherent high dimensionality implies an expansive design space that can be readily accessed to elicit a varied range of behaviours from granular jamming actuators.

The most popular type of granular jamming actuator, the bag gripper~\cite{brown2010universal,amend2012positive}, comprises a roughly spherical flexible membrane housing numerous grains (e.g., ground coffee), and attached to a vacuum line.  At standard pressure, the grains possesses a fluid-like quality, allowing the gripper to be pushed against a target object and deform around it.  Subsequently pulling a vacuum causes the granular material to solidify, increasing inter-grain compressive and frictional forces and causing a grip.  

Tuning constituent grain properties is an under-explored research area.  To date, studies either only test a single grain type, only consider one free variable, e.g., grain size, with other factors held constant, or use a handful of commercially-available grain types (e.g., \cite{chopra2020granular}), which limits testable grains to a somewhat arbitrary selection of those that can be procured, as well as offering no guarantees over the geometric accuracy of the grain. Typical grains include the aforementioned coffee grounds, as well as rice, glass spheres, plastic spheres, and rubber cubes.  None of these approaches allow for grain shape and size to be precisely controlled and rigorously explored in a principled manner.

3D printing is a promising avenue to realise this goal, with an recent study demonstrating the in-principle ability to tune performance properties for jamming actuators~\cite{An_2020}, in this case using only spheres, but varying levels of filleting. Considering the wide range of grain properties which can be precisely specified and fabricated, and the limited explorations of this topic to date, we are motivated to experimentally verify the benefit 3D printing as an avenue towards creating bespoke, high-performance jamming actuators.  

In this paper we explore the impact of 3D printed grains of different shapes and sizes on gripping performance.  We first design a set of grain shapes that are predicted by DEM simulations to provide a diverse array of gripping behaviours.  We then 3D print those grains in four different sizes to provide a comprehensive test set.  Each shape/size combination is used as the constituent grain in a simple bag-style jamming gripper, and gripping performance is assessed across four test objects via pull-off testing.  In addition, we quantify the accuracy of 3D printing as a fabrication technique.  A Confocal Laser Microscope is used to characterise the anisotropic roughness of the printed grains and compare to the idealised CAD model.  We show good shape accuracy and provide some guidelines around suitable grain sizes to use, which is critically important when addressing reality gap issues (an area where granular jamming grippers are particularly vulnerable) when, e.g., linking experimental studies to DEM modelling.

We summarise the key novel contributions of this work as follows:

\begin{itemize}
    \item The first comprehensive study of shape and size variation in 3D printed granular jamming grippers, with testing on an applied gripping task.  This includes use of a comprehensive test set of 12 grains (3 sizes of 4 grain shapes; Fig.~\ref{fig:fig1}), with performance assessed on 4 test objects,  
    \item The first demonstration of the superquadric grain formulation to specify grain shape in a soft robotics context, and,
    \item The first characterisation of 3D printed grains in terms of shape-accuracy.
\end{itemize}

Results demonstrate that grain shape and size are key determinants in granular gripping performance, and control over both is a key (although currently heavily underutilised) technique to provide  high-performance, bespoke actuation behaviour.  Moreover, we demonstrate that there is no universally-optimal grain shape for gripping, and that gripper performance with varying size and shape is often unintuitive.  This suggests the use of computational optimisation techniques as a promising path towards fully harnessing particle size and shape to expand the range and performance of granular grippers across a range of diverse usage scenarios


\begin{figure}[t!]
\centering
\includegraphics[width=0.8\columnwidth]{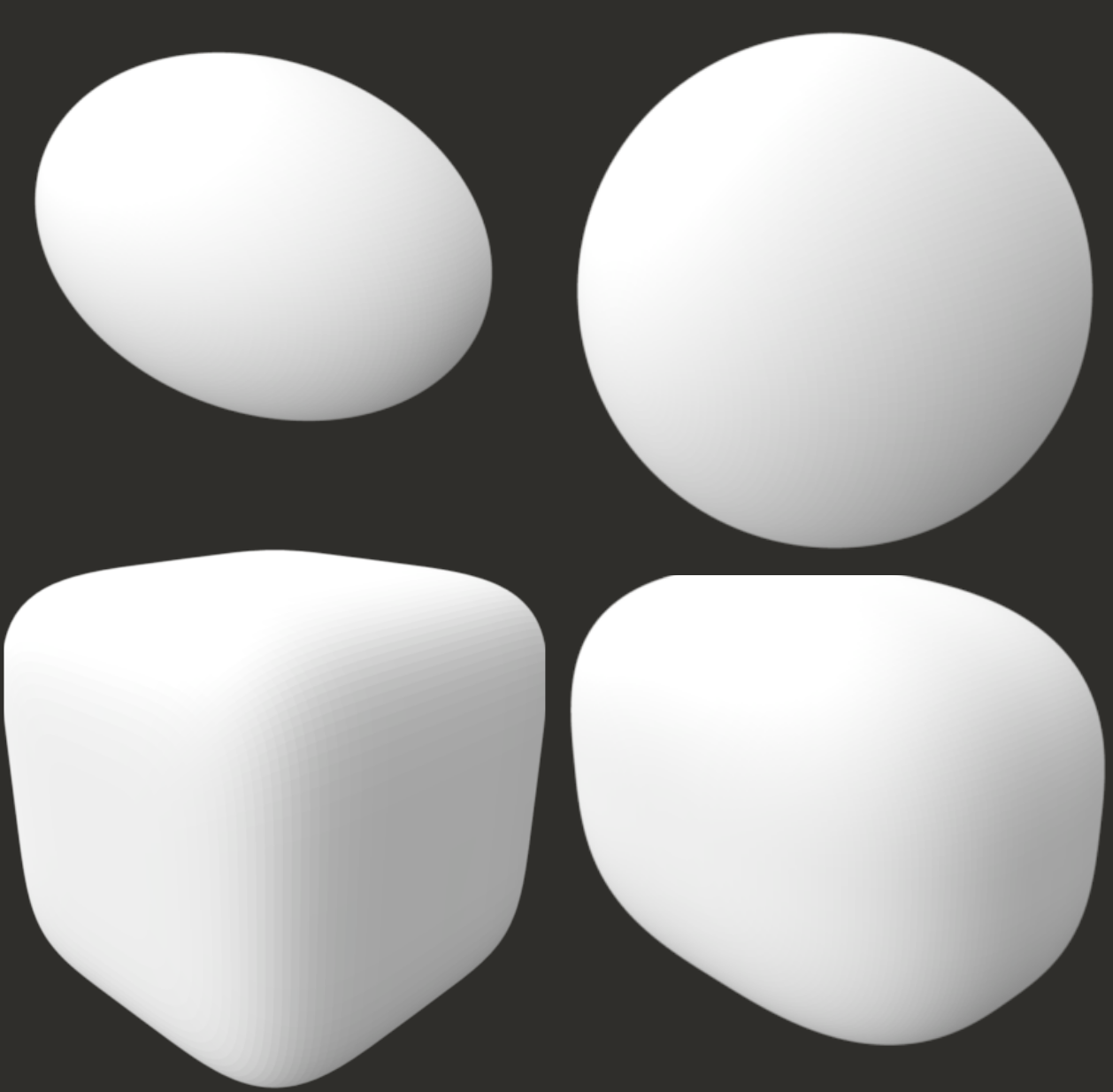}
\caption{CAD images of the four ideal superquadric grains, clockwise from top-left: ellipsoid, sphere, superellipsoid, superball.}
\label{fig:fig2}
\end{figure}

 \section{BACKGROUND}
\label{sec:related_work}
Here we detail two relevant areas of literature, (i) Jamming in soft robotics, and (ii) studies on grain shape.

\subsection{Jamming soft robotics}

Jamming occupies a niche in soft actuation where comparatively rapid response times ($\approx$1s) and large stiffness variations are required~\cite{shintake2018soft}.  Jamming can be categorised into three types depending on the mechanism that induces stiffness variation - granular jamming (inter-particle forces on compression), layer jamming~\cite{narang2018mechanically} (forces between stacked layers of material), and fibre jamming~\cite{brancadoro2018preliminary}(interactions between fibres).  The three mechanisms have different strengths and weaknesses; see~\cite{jamming_review} for a recent comprehensive review.

Granular jamming is the most prevalent, with a diverse array of actuator configurations allowing for locomotion of worm-like and snake-like robots~\cite{steltz2010jamming,Robertsoneaan6357}, soft robotic 'paws'~\cite{hauser2018compliant,chopra2020granular}, and minimally-invasive surgery tools~\cite{cianchetti2013stiff}. 

The literature evidences an increasing variety of gripper designs, from fingers and hands to hybrid jamming/pneumatic actuators~\cite{wei2016novel},  primarily facilitated through 3D printing of moulds for subsequent casting, or more recently direct 3D printing~\cite{zhu2019fully,stano2021}.  An increasing range of actuation techniques are similarly observed, including positive pressure~\cite{amend2012positive} and fluidisation~\cite{kapadia2012design}, to jamming based on geometric confinement~\cite{li2017passive,li2018distributed} (e.g. by pressurising and expanding a neighbouring cavity), in addition to the standard negative pressure technique.

Although these more complex designs are having a substantial impact in their relevant domains, the most popular variant remains the bag-style 'universal' gripper~\cite{brown2010universal,amend2016soft}, which has been combined with learning for object-specific gripping strategies~\cite{jiang2012learning} and found use in diverse range of application areas, including sub-sea sampling~\cite{licht2017stronger}, and prosthetic end-effectors~\cite{cheng2016prosthetic}, .

\subsection{Custom grains}
Naturally-occurring grains such as coffee have implicit uncontrolled variations in shape and size distribution.  Control over shape allows for bespoke grain properties, and can be realised straightforwardly through modelling techniques (predominantly DEM), and more recently via 3D printing.

Grain shape determines fundamental bulk properties including stiffness and stress responses~\cite{athanassiadis2014particle}, as well as packing fraction~\cite{delaney2010packing}.   Jamming structures based on a bonded sphere representation have been optimised via genetic algorithm ~\cite{miskin2013adapting,jaeger2015celebrating}, with evolution able to create a small library of grains across the continuum of the fundamental variable under consideration.  Grains were printed for experimental confirmation of results, although some instantiations of the representation were unprintable. Later work used multi-objective evolution to explore a range of bulk properties, including packing fraction, with represented by parameterised superquadrics~\cite{RN406}.  Superquadrics are inherently printable, and thus a strong candidate representation.  We note that all of the above work is carried out in the context of granular physics, and lacks direct applicability to soft robotics.   

In a robotics context, information on grain performance is more scarce.  We know that performance is task-dependent - work on compliant paws, for example, tends to favour cubes~\cite{hauser2018compliant} due to their tendency to geometrically jam. Damping effects of variations of 3D printed grains with various filleting patterns highlights the flexibility of 3D printing to precisely control grain fabrication, in this case for vibration damping, however  only spheres were considered~\cite{An_2020}.

\subsection{Literature summary}

Although the impact of membrane material in bag-style grippers has been studied in-depth~\cite{jiang2014robotic}, there is no comprehensive study covering applied testing of shape effects of 3D printed grains for jamming actuators.  We are therefore motivated to study a diverse range of grain shapes and sizes in the context of jamming grippers, where we can easily study and precisely fabricate any desired shape.   So far, only a very small subset of possible grain geometries (spheres and cubes) have been studied for soft robotics applications. 

We specifically note the novelty and applicability of the superquadric formulation in this context for generating guaranteed-printable grains. Moreover, the superquadric formulation allows access to a range of ellipsoidal grain types which have found significant uptake in other fields \cite{delaney2010packing}, and may be straightforwardly transferable to the soft robotics domain.  These shapes are not easy to purchase and typically require printing.

 \section{MATERIALS AND METHODS}
\label{sec:methodology}

We aim to show how gripper performance, even in simple 'universal' bag grippers, is critically dependent on the containing grains, and furthermore how 3D printing facilitates bespoke jamming performance for specific gripping applications.  To make results more broadly applicable, we use the most popular bag-style gripper, and a standard measurement of performance (pull-off force). 

\subsection{Grain design and fabrication}
We chose four different grain shapes from the family of possible superellipsoids, that have been shown from DEM simulations to maximise the diversity of behaviours of the granular material in the jammed state~\cite{delaney2010packing,10.1145/3377929.3389951} (Fig.~\ref{fig:fig2}).  Grains are parameterised following:

\begin{equation}
  (x/a)^{m} + (y/b)^{m} + (z/c)^{m} = 1.
\end{equation}

The first shape is a {\it sphere} ($m$=2, $a$=$b$=$c$=1), commonly employed in both simulation and experiments. Spheres lack the additional degrees of freedom of a non-spherical shape and have a highly distinct jamming density and lower average contact number compared to other superellipsoidal shapes. To incorporate the first anisotropic effects of aspect ratio, we use a prolate {\it ellipsoid} with two equal axes ($m$=2, $a$=1, $b$=$c$=0.65). This shape introduces a rotational degree of freedom and a significantly higher jamming packing density and average contact number compared to spheres. 

Flat surfaces have a strong impact on the degree of orientational ordering that readily forms within a jammed system and creates unique face-on-face contacts that strongly oppose both rotation and lateral movement. To study this effect we utilise an equiaxed {\it superball} with a shape factor (exponent in the standard superellipsoid formula) of $m$=5, $a$=$b$=$c$=1. This shape is similar to a cube with rounded corners and can readily produce jammed packings with high degrees of orientational ordering along the 3 primary axes of the grains (cubatic ordering). Finally we select a triaxial {\it superellipsoid} with a shape factor of $m=3$ and aspect ratios of $a$=1 $b$=0.75 $c$=0.6, giving a cuboidal shape with highly smoothed corners. This shape sits at the intermediate points of surface curvature and angularity, with three distinct aspect ratios designed to provide a moderate amount of face-on-face contacts, while disrupting the formation of long-range orientational ordering. This set includes two popular shapes in the literature (spheres and cubes (superballs)), and two that are highly underexplored (ellipsoids and superellipsoids).
 
Grains are printed on the Stratasys Connex3 OBJET500 polyjet printer in the rigid vero material at 16 micron layer height.  Printing time was independent of grain shape but varied slightly due to grain size; 1000 3mm grains took $\approx$ 53 minutes while 1000 7mm grains took $\approx$ 1 hour. Once printed, the grains were placed into a 1mm aperture laboratory sieve and placed in a Stratasys CSIIP CleanStation support removal bath for 24 hours.  For consistency, grains at each quoted "size" have equivalent volume to a sphere of that size, which we call Sphere-Volume Equivalent (SVE).  For example, a 5SVE superball is a superball scaled such that its volume is equal to a 5mm diameter sphere.

\begin{figure*}[t]
\centering

\subfloat{\includegraphics[height=5.5cm,width=3cm]{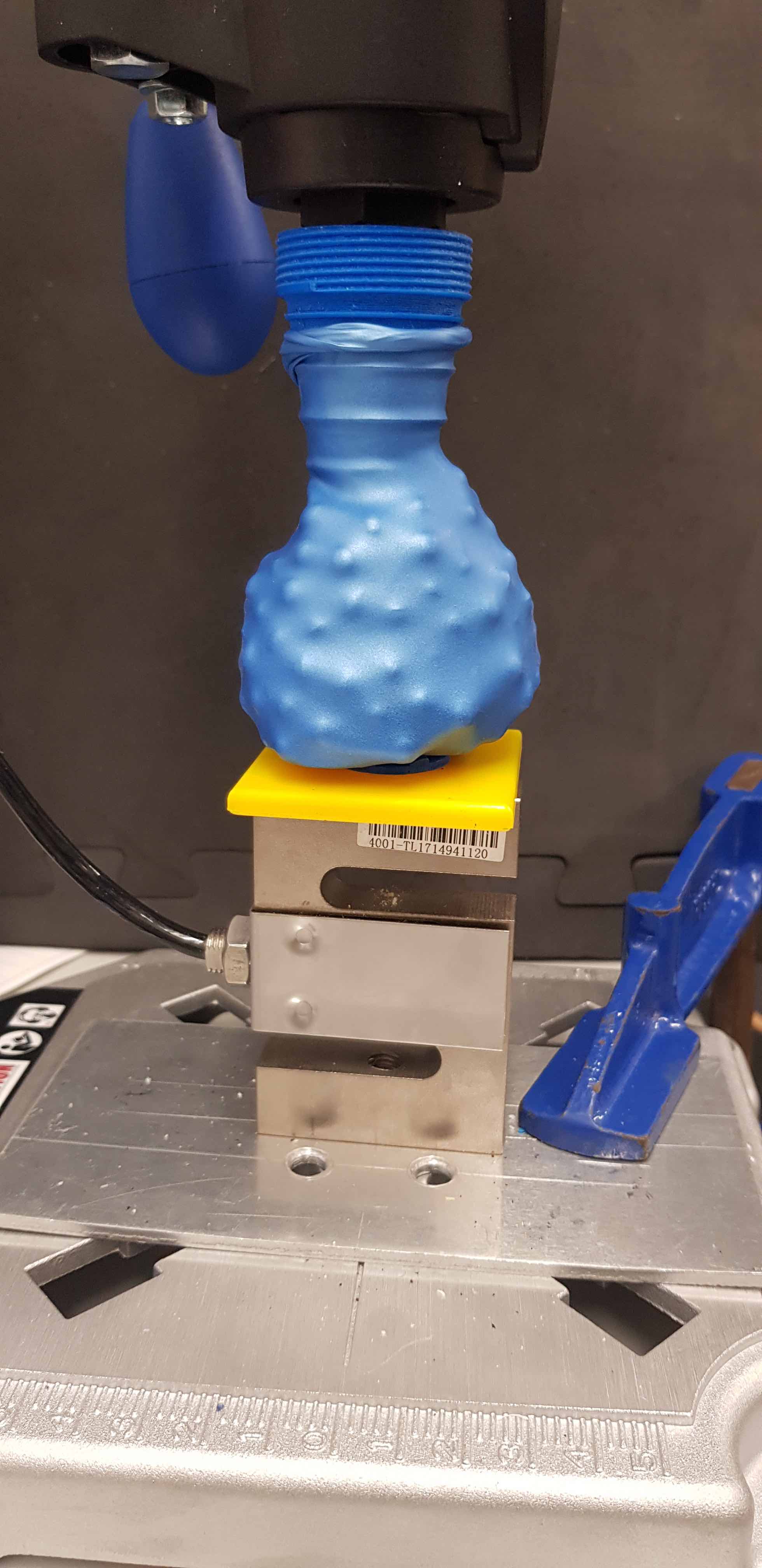} }
\subfloat{\includegraphics[height=5.5cm,width=3cm]{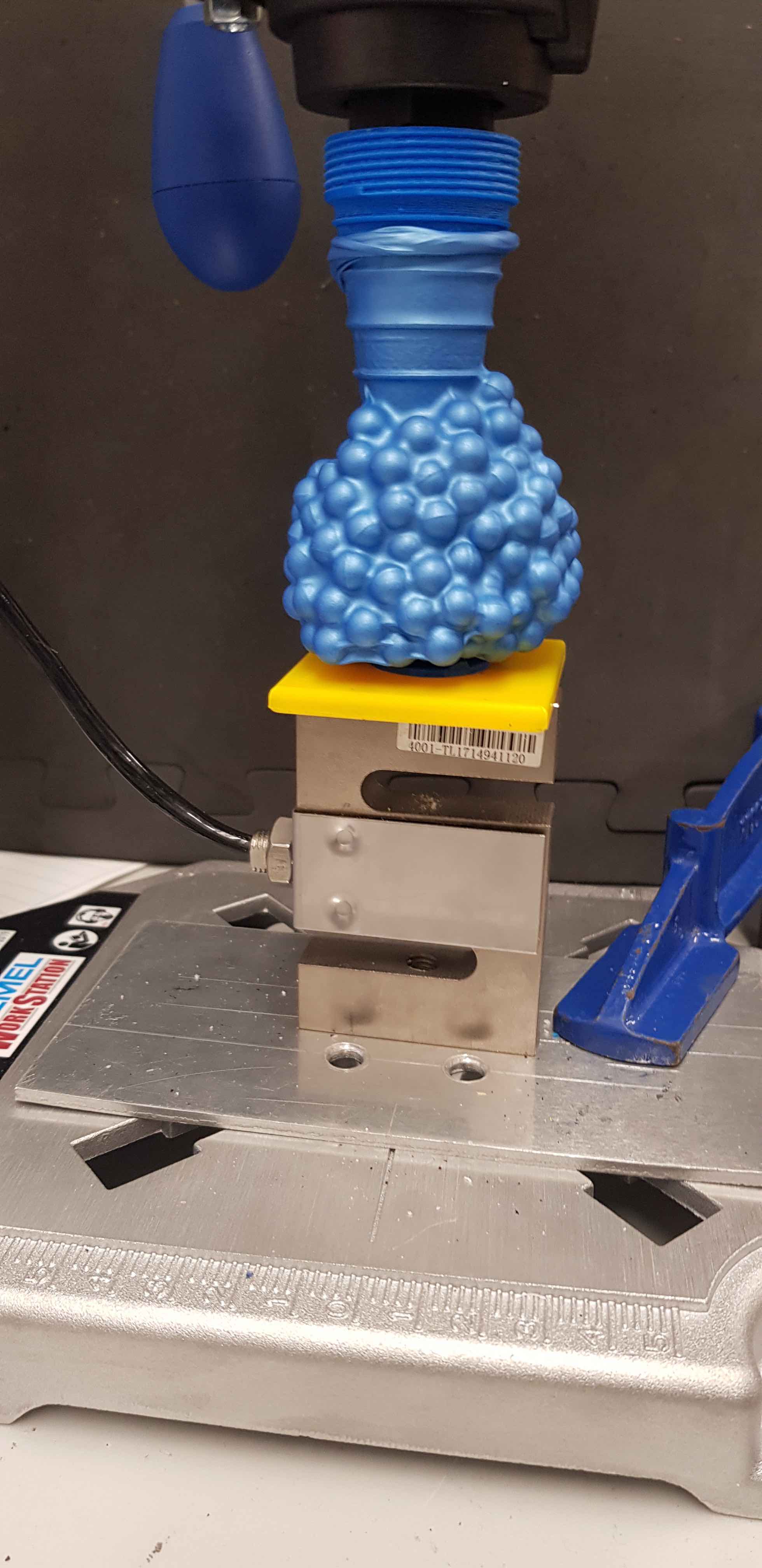} }
\subfloat{\includegraphics[height=5.5cm,width=3cm]{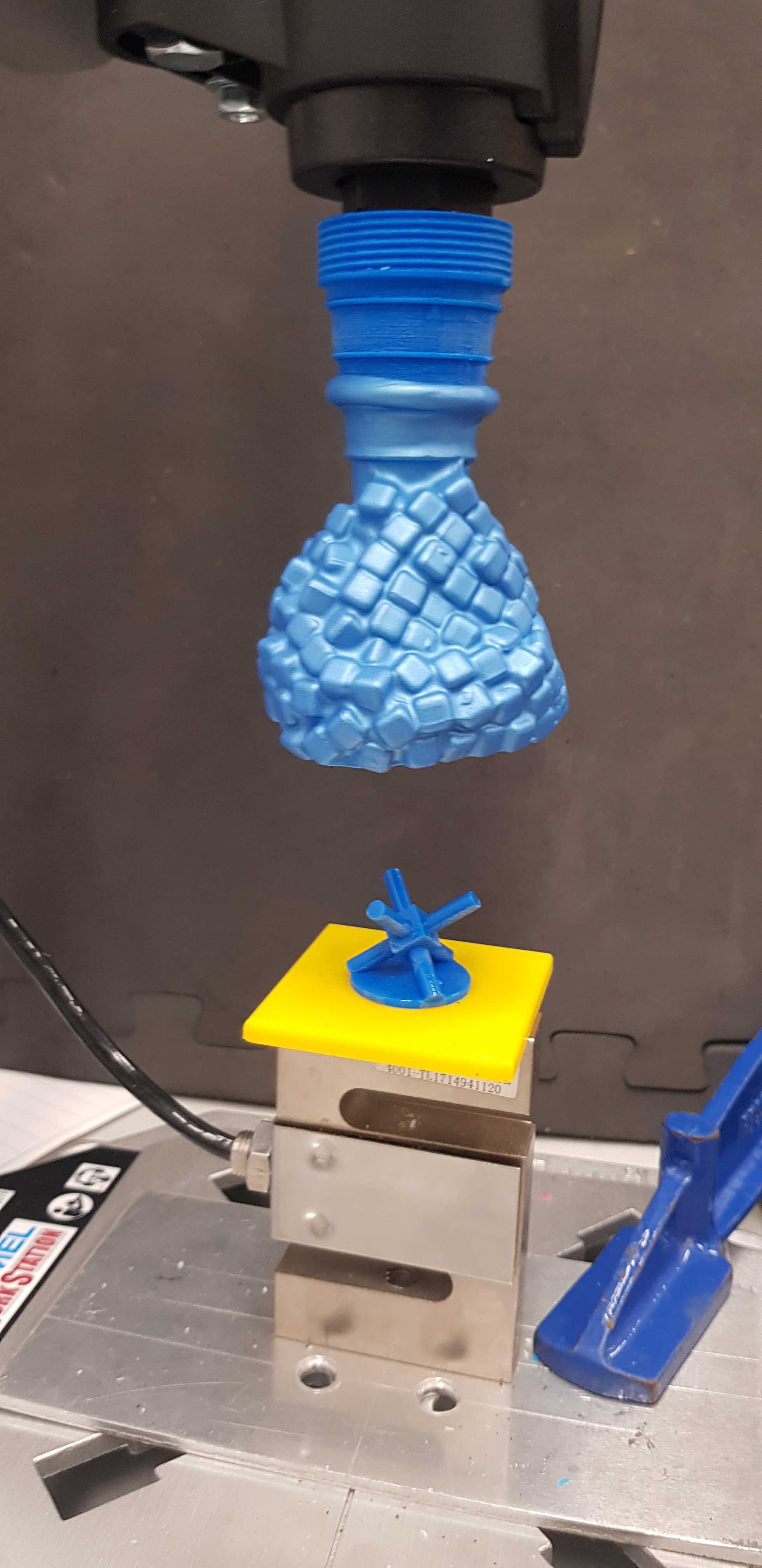} }
\subfloat{\includegraphics[height=5.5cm,width=3cm]{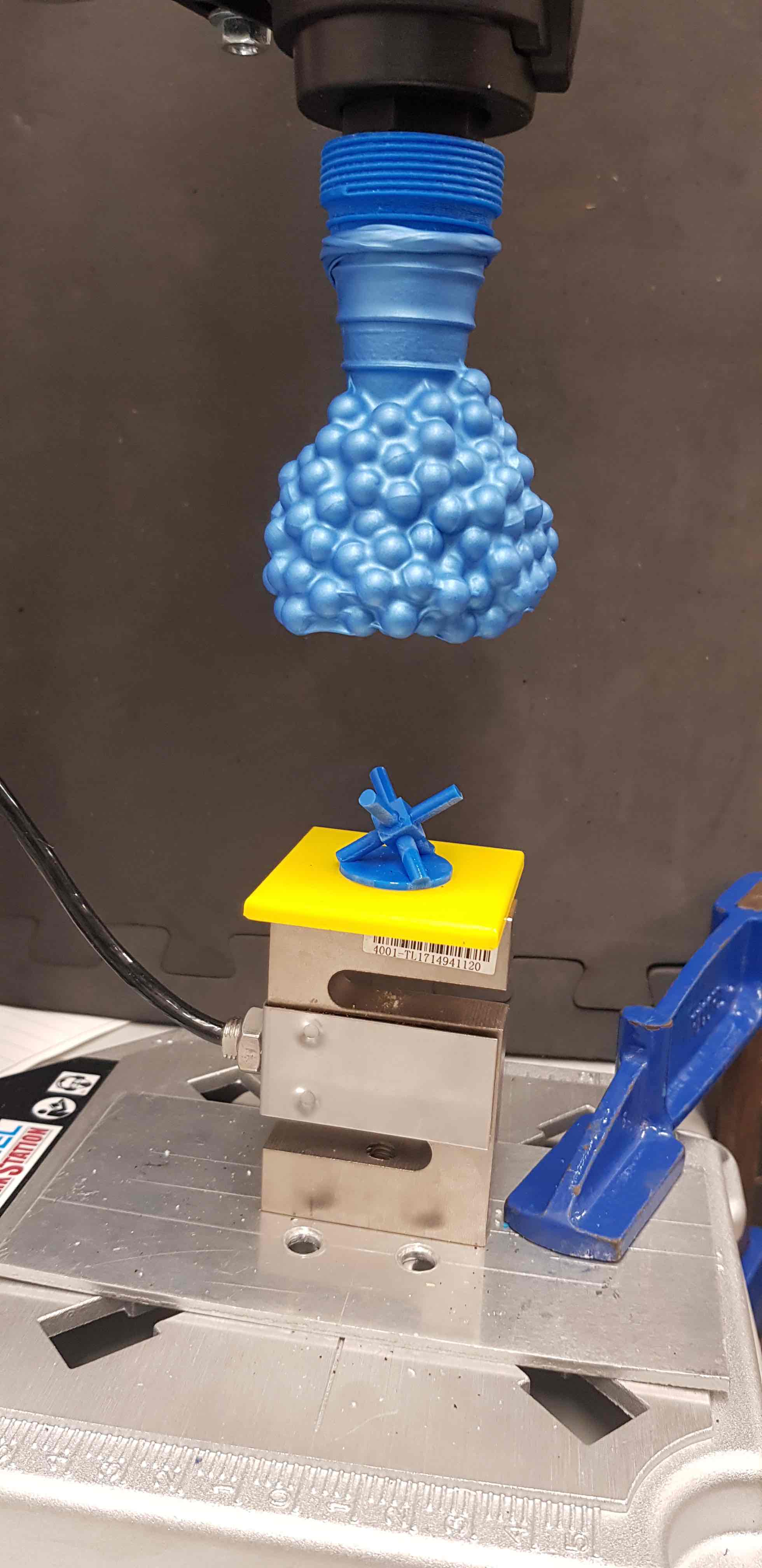} }

\caption[]{Showing the testing process with 7SVE spheres.  A load cell is clamped to a desk, with a yellow plastic plate providing a horizontal plane for gripping.  The test object (star) is screwed into the load cell.  The gripper is attached to a drill press stand to constrain range of motion to vertical only. CW from top left: gripper lowered onto object (unjammed); jammed on object; jammed and pulled from object; 7SVE superballs jammed and pulled from object (shown for comparison with geometric jamming evident).}
\label{fig:fig4}
\end{figure*}

\begin{figure*}[ht!]
\begin{center}
\centering 
 \subfloat(a){\includegraphics[width=8cm,height=6.5cm]{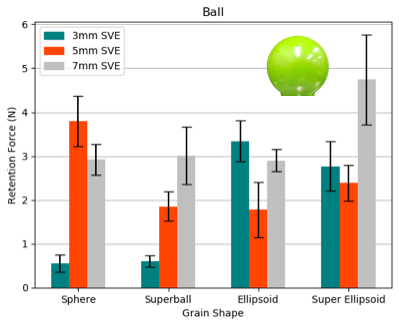} }
 \subfloat(b){\includegraphics[width=8cm,height=6.5cm]{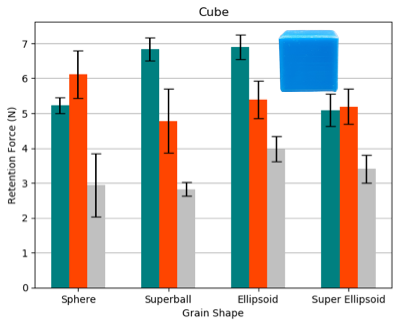} }\\
 \subfloat(c){\includegraphics[width=8cm,height=6.5cm]{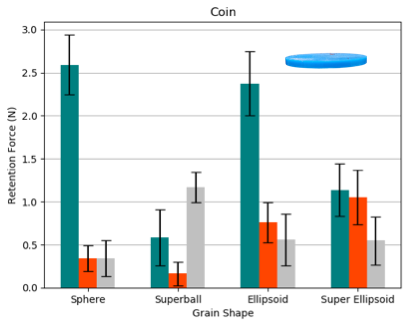} }
 \subfloat(d){\includegraphics[width=8cm,height=6.5cm]{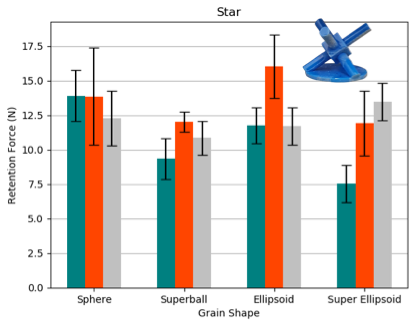} }
\end{center}
\caption[]{Results for the mean pull-off force (N), averaged over 5 repeats,  on the four test objects; (a) ball, (b) cube, (c) coin, (d) star.  Error bars denote standard error. Colour indicates grain size. Images of respective test objects are provided with each sub figure.}
\label{fig:fig3}
\end{figure*}

\subsection{Experimental setup}

A thin, 12.5cm latex balloon was chosen to reduce effects of the membrane on performance.  The balloon was marked with a horizontal line, just below the neck, and filled up to the mark with the chosen grain type using a standard funnel. The balloon was attached to a 3D printed adapter and mounted onto a drill press stand to constrain the range of motion to vertical only, aiding repeatability. A Thomas 107CDC20-H vacuum pump was connected to the balloon gripper via 5.5mm silicone tubing. 

Four test objects (a ball, cube, coin, and star) were chosen to provide diverse gripping challenges whilst being easily integrated into the test rig. The test objects were all of approximately 20mm$^3$ in size, chosen so as to be in the range of 3x to 7x the size of the particles used in the gripper so as to maximise the size and shape variations measured in the experiment. Test objects were 3D printed with a thread and screwed into a Zemic H3-C3 load cell, which was clamped and centered inline with the balloon adapter. A flat platform (50mm x 50mm) was attached between the object and load cell to replicate the action of picking an object from a flat surface. 

Testing involved lowering the gripper onto the object in an unjammed state. The vacuum was activated to cause jamming, and the gripper slowly raised until it released and completely cleared the test object. The vacuum was deactivated and the gripper manually reset by shaking. Fig.~\ref{fig:fig4} the testing procedure.  Each grain shape/size combination was tested 5 times per test object and mean performance reported.

\section{RESULTS}
Results are displayed in Table~\ref{table:tab1} and Fig.~\ref{fig:fig3}. Statistical significance is assessed via a Mann-Whitney U-test with a p-value $<$0.05. 

\begin{table}[t!]
    \caption{  Pull-off testing ($N$) data for all grain shape and size combinations, across the four test objects. Results are averaged over 5 repeats, with standard error in parenthesis.}
    \centering
    \begin{tabular}{p{1.175cm}p{1.3cm}p{1.3cm}p{1.3cm}p{1.3cm}}
    \toprule
       Object  & Sphere   & Superball  & Ellipsoid & Super ellipsoid\\
    \midrule
            3mm ball &  0.55 (0.2)     & 0.60 (0.1)  &3.34 (0.5)   &2.77 (0.6) \\ 
            5mm ball & 3.80 (0.6)     & 1.85 (0.3)   & 1.78 (0.6)   &2.38 (0.4)\\
            7mm ball &  2.92 (0.4)     & 3.00 (0.7)   & 2.90 (0.3)   &4.74 (1.0)\\  
             \midrule          
            3mm cube & 5.23 (0.2)  &6.84 (0.3)& 6.90 (0.4)& 5.09 (0.5)\\
            5mm cube& 6.10 (0.7)  &4.78 (0.9)&5.39 (0.5)& 5.19 (0.5)\\
            7mm cube& 2.94 (0.9)  &2.82 (0.2)&3.98 (0.4)& 3.41 (0.4)\\
             \midrule           
            3mm coin& 2.59 (0.4)  &0.58 (0.3)& 2.38 (0.4)& 1.14 (0.31)\\
            5mm coin& 0.34 (0.2)  &0.16 (0.1)& 0.76 (0.2)& 1.05 (0.3)\\
            7mm coin&  0.34 (0.2)  &1.17 (0.2)& 0.56 (0.3)& 0.55 (0.3)\\
            \midrule         
            3mm star& 13.90 (1.8)  &9.34 (1.5)& 11.75 (1.3)& 7.52 (1.4)\\
            5mm star& 13.85 (3.5)  &12.02 (0.7)& 16.03 (2.3)& 11.92 (2.4)\\
            7mm star& 12.28 (1.9)  &10.84 (1.2)&11.68 (1.4)& 13.48 (1.6)\\
    \bottomrule
    \end{tabular}
    \label{table:tab1}
\end{table}

\subsection{Size Effects}
For the test objects used, we see a general trend of difficulty: coin is most difficult (0.967N mean across all tests) (Fig.~\ref{fig:fig3}(c)), followed by ball (2.553N) (Fig.~\ref{fig:fig3}(a)), cube (4.889N) (Fig.~\ref{fig:fig3}(b)), and star (12.05N) (Fig.~\ref{fig:fig3}(d)).  It is known from the literature that coins are difficult targets for bag-style jamming grippers~\cite{amend2012positive}. Interestingly, we see 3SVE grains (sphere and ellipsoid) allow for reasonable coin grips without use of fluidization, with a very large drop off in the gripping strength for the 5mm and 7mm grains. This may be due to the key dimension of the coin (it's height of 2.5 mm) compared to the grain axis length, where for small spheres/ellipsoids the greater ability for the grains to flow and surround the coin improves the gripping strength, while for larger grains the coin effectively sits below their half axis length, greatly reducing gripping strength. 

For the cube object, there is a general trend of decreasing grip strength with increasing particle size across all particle shapes, with quite consistent gripping strengths across the different particle shapes. This consistency is likely due to the simplicity of the cube shape, with the grip being primarily driven by contacts along a set of perfectly flat planes with variations seen being being due to the greater ability of the smaller particles to conform to the cube faces. 

The star is a particularly interesting shape owing to its non-convexity, with spikes that can become embedded within the jammed gripper structure, allowing it to be supported in multiple places from below. This creates significantly higher gripping strengths for the star compared to the other objects across all particle shapes and sizes, with the complexity of the interactions of the spikes with the individual particles (which have a similar length scale) leading to strong non-linear variations in the gripping performance over the set of particles considered. 

The ball object presents a uniformly curved surface to the gripper that can be particularly difficult to grip in cases where the gripper fails to deform around the shape below it's medial axis. The complexity of interaction with the curved surface is evident in the data, with high variability in the gripping performance found as particle size and shape is varied. 


\subsection{Shape Effects}
The ellipsoid is the best performing grain across all test objects and sizes (5.619N), then sphere (5.404N), superellipsoid (4.936N), and superball (4.5N). This suggests increased gripping performance for the target objects considered where the particles have highly curved surfaces (spheres and ellipsoids) vs a more a angular shape with flatter faces (superballs and superllipsoids). The curved surfaces permit the particles to flow more easily around the target object, allowing the gripper to better deform around the object and achieve a tighter grip. The increased gripping strength of the ellipsoid vs the sphere and the superellipsoid vs the superball also suggests that a degree of anisotropy in the particle shape increases the grip strength, which could be attributable to the higher density and greater number of particle contacts that these grain shapes have in the jammed state \cite{delaney2010packing}.

The data clearly demonstrates a strong interaction effect between the target object shape, the particle shapes within the gripper and the particle sizes, showing clearly that there is no single optimal shape for gripping a given object across all particle sizes. It is interesting to note that the more unconventional particle shapes (ellipsoids and superellipsoids) have superior performance to the more traditional shapes (spheres and cubes) in several cases; a promising indication that further exploration may yield even more interesting shape-performance combinations.  



\begin{figure*}[t!]
\includegraphics[width=0.9\textwidth]{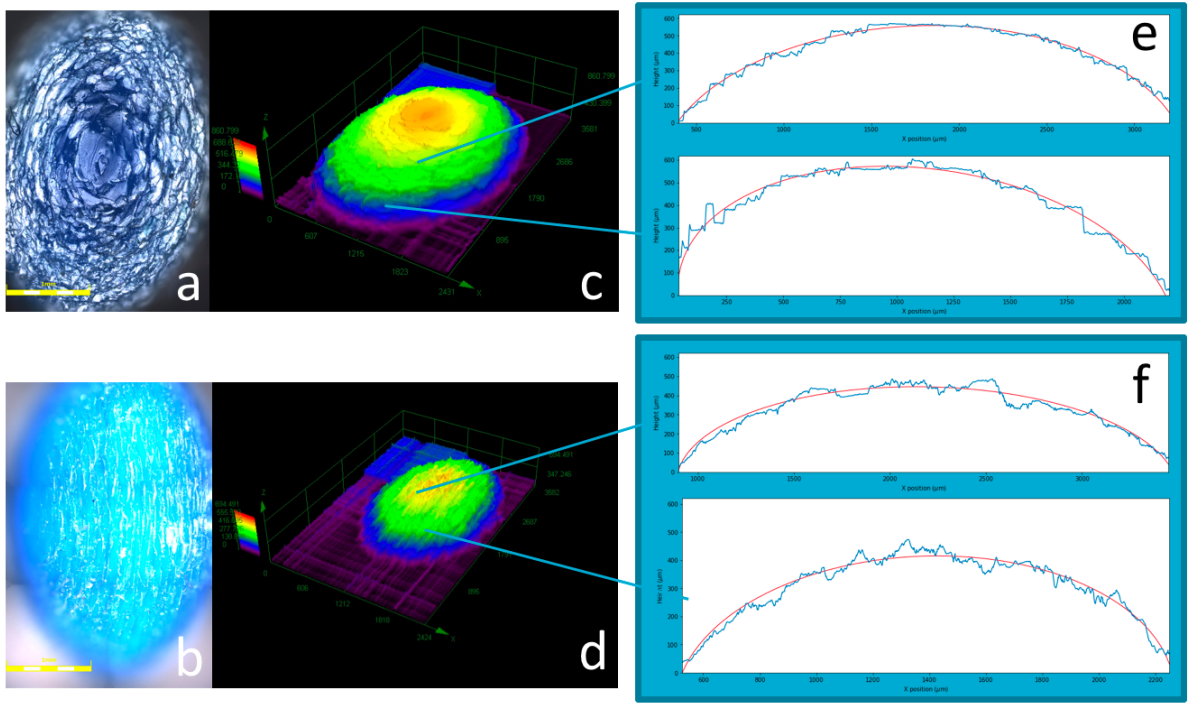}
\centering
\caption[]{Results from Confocal Laser Scanning Microscope imaging of a 3mm ellipsoid. Left Column: optical images of (a) top view (b) side view, with layers clearly visible.  Middle Column: height maps for (c) top view and (d) side view. Right Column (e) and (f): line profiles across the surfaces of the ellipsoid and the corresponding ideal 2D elliptical line at that location.}
\label{fig:fig5}
\end{figure*}
  
\section{SHAPE-NOISE APPROXIMATION}
To assess the degree to which our 3D printing fabrication method is able to accurately represent a particular grain shape, we imaged the surfaces of the grains using an Olympus LEXT OLS4100 Confocal Laser Scanning Microscope with a scan resolution of 0.01$\mu m$. Optical images and surface laser scans of the surface of an ellipsoid with a primary axis length of 3mm and aspect ratios for the other two axes of 0.65 is shown in Fig.~\ref{fig:fig5} (a)-(d). The polyjet printer used has a layer resolution of 16 microns, and visual inspection of Fig.~\ref{fig:fig5}(a)-(b) clearly shows the resulting surface roughness and anisotropy on the grain surfaces. 

A series of line profiles along the surface of the ellipsoid were obtained and analysed to quantify the deviation of the printed surface from the ideal ellipsoid shape (see Fig.\ref{fig:fig5} (e)-(f)). The 2D line profiles should each be exactly represented as an ellipse if the 3D ellipsoid shape were being perfectly reproduced, however due to the finite print resolution this will not be the case. We analysed the deviation away from the idealised ellipse (obtained by fitting an elliptical function to the data) for two sets of line profiles, one measured on the surface produced directly under the print head and the other in the orthogonal direction. We quantified the surface roughness using both the average maximal variation of the surface away from the ideal line profile, $R_{max}$, and the average root mean squared deviation, $R_q$. For the surface directly under the print head we measured $R_q = 29\mu m$ with maximal deviations up to $R_{max} = 80 \mu m$, while for the orthogonal direction we measured a higher $R_q = 41 \mu m$ with similar maximal deviations up to $R_{max} = 80 \mu m$. These surface measurements demonstrate the very good fidelity with which we can reproduce a specific grain shape, even for grains that are only a few mm in size. The shape approximation error will vary with the size of the grain, with for example a 3 mm sphere having average deviations of its surface from its idealised shape of about 1.2 \% of the sphere diameter, with maximum deviations around 3 \%. As the particle size is increased the percentage error reduces proportionally, with a 7 mm particle having average deviations of about 0.5 \% of the sphere diameter.  Choice of a suitable grain size is tied to both (i) dependence on precise shape information to produce the desired behaviour, and (ii) whether or not DEM modelling is used.  DEM uses idealised shape representations, so it is likely that larger sized grains with comparatively smaller shape deviations would act to reduce reality gap effects.


 \section{DISCUSSION}
\label{sec:discussion}

In this paper we analysed size and shape effects of 3D printed granular jamming grippers, demonstrating that 3D printing is a simple, direct method to realise controllable shape variations.  Results show that underexplored grain shapes (ellipsoids, superellipsoids) are at least competitive with the more popular spheres and cubes, with ellipsoids in particular performing strongly across all test objects.  Given the general lack of diversity in the literature regarding grain shape, it follows that many granular actuators could achieve heightened performance through grain shape optimisation, and that further research into programmable behaviour through grain shape tuning is indicated.  

Aside from a clear display of relative object difficulty, results display few clear trends. This is unsurprising, as the behavioural properties of granular media are strongly dependent on the constituent particle properties and the geometry of objects with which they interact, and speaks to the wide behavioural envelope that grain design allows us to access.   Coupling 3D printing, DEM, and physical experimentation is an exciting research avenue for expanding knowledge of granular physics, where modelling could be easily verified with physical grain experimentation with intrinsic high shape accuracy.  This may also include tailoring granular material properties, e.g., varying Shore harness, which is simple to realise with polyjet printing.  Practically, printing grains paves the way towards bespoke granular actuators, with future research in multiscale optimisation from grains to full robots in a 'materials to machines' approach~\cite{RN414}.

We also demonstrated the power of superquadric grain representations in a soft robotics context to diversify gripper performance, even with relatively constrained parameter choices.  Expanding the parameter bounds would open up more unconventional grains and potentially higher performance, increasing scope to tune jamming behaviour.

Additionally an analysis of the surface roughness of our 3D printed particles comparing to the ideal surface shape shows the ability of 3D printing to produce fine details and sufficient shape representation even for particles of only a few mm in size.  We position this work as the start of an attempt to build a body of knowledge to facilitate practitioners in selecting optimal grains in a principled, application-dependent manner and pave the way towards a new generation of bespoke jamming actuators.  
  
\section*{ACKNOWLEDGEMENT}
This work was funded by CSIRO's Active Integrated Matter Future Science Platform.  We thank Mark Greaves and Julian Ratcliffe for generating the roughness profiles.


\bibliographystyle{IEEEtran}
\bibliography{jamming,jamming2,jamming3}

\begin{thebibliography}{10}
\providecommand{\url}[1]{#1}
\csname url@rmstyle\endcsname
\providecommand{\newblock}{\relax}
\providecommand{\bibinfo}[2]{#2}
\providecommand\BIBentrySTDinterwordspacing{\spaceskip=0pt\relax}
\providecommand\BIBentryALTinterwordstretchfactor{4}
\providecommand\BIBentryALTinterwordspacing{\spaceskip=\fontdimen2\font plus
\BIBentryALTinterwordstretchfactor\fontdimen3\font minus
  \fontdimen4\font\relax}
\providecommand\BIBforeignlanguage[2]{{%
\expandafter\ifx\csname l@#1\endcsname\relax
\typeout{** WARNING: IEEEtran.bst: No hyphenation pattern has been}%
\typeout{** loaded for the language `#1'. Using the pattern for}%
\typeout{** the default language instead.}%
\else
\language=\csname l@#1\endcsname
\fi
#2}}

\bibitem{hughes2016soft}
J.~Hughes, U.~Culha, F.~Giardina, F.~Guenther, A.~Rosendo, and F.~Iida, ``Soft
  manipulators and grippers: a review,'' \emph{Frontiers in Robotics and AI},
  vol.~3, p.~69, 2016.

\bibitem{shintake2018soft}
J.~Shintake, V.~Cacucciolo, D.~Floreano, and H.~Shea, ``Soft robotic
  grippers,'' \emph{Advanced Materials}, vol.~30, no.~29, p. 1707035, 2018.

\bibitem{jamming_review}
S.~Fitzgerald, G.~Delaney, and D.~Howard, ``A review of jamming actuation in
  soft robotics,'' \emph{Actuators}, vol.~9, no.~4, p. 104, 2020.

\bibitem{brown2010universal}
E.~Brown, N.~Rodenberg, J.~Amend, A.~Mozeika, E.~Steltz, M.~R. Zakin,
  H.~Lipson, and H.~M. Jaeger, ``Universal robotic gripper based on the jamming
  of granular material,'' \emph{Proceedings of the National Academy of
  Sciences}, vol. 107, no.~44, pp. 18\,809--18\,814, 2010.

\bibitem{amend2012positive}
J.~R. Amend, E.~Brown, N.~Rodenberg, H.~M. Jaeger, and H.~Lipson, ``A positive
  pressure universal gripper based on the jamming of granular material,''
  \emph{IEEE Transactions on Robotics}, vol.~28, no.~2, pp. 341--350, 2012.

\bibitem{chopra2020granular}
S.~Chopra, M.~T. Tolley, and N.~Gravish, ``Granular jamming feet enable
  improved foot-ground interactions for robot mobility on deformable ground,''
  \emph{IEEE Robotics and Automation Letters}, 2020.

\bibitem{An_2020}
S.-Q. An, H.-L. Zou, Z.-C. Deng, and D.-Y. Guo, ``Damping effect of
  particle-jamming structure for soft actuators with 3d-printed particles,''
  \emph{Smart Materials and Structures}, vol.~29, no.~9, p. 095012, aug 2020.

\bibitem{narang2018mechanically}
Y.~S. Narang, J.~J. Vlassak, and R.~D. Howe, ``Mechanically versatile soft
  machines through laminar jamming,'' \emph{Advanced Functional Materials},
  vol.~28, no.~17, p. 1707136, 2018.

\bibitem{brancadoro2018preliminary}
M.~Brancadoro, M.~Manti, S.~Tognarelli, and M.~Cianchetti, ``Preliminary
  experimental study on variable stiffness structures based on fiber jamming
  for soft robots,'' in \emph{2018 IEEE International Conference on Soft
  Robotics (RoboSoft)}.\hskip 1em plus 0.5em minus 0.4em\relax IEEE, 2018, pp.
  258--263.

\bibitem{steltz2010jamming}
E.~Steltz, A.~Mozeika, J.~Rembisz, N.~Corson, and H.~Jaeger, ``Jamming as an
  enabling technology for soft robotics,'' in \emph{Electroactive Polymer
  Actuators and Devices (EAPAD) 2010}, vol. 7642.\hskip 1em plus 0.5em minus
  0.4em\relax International Society for Optics and Photonics, 2010, p. 764225.

\bibitem{Robertsoneaan6357}
\BIBentryALTinterwordspacing
M.~A. Robertson and J.~Paik, ``New soft robots really suck: Vacuum-powered
  systems empower diverse capabilities,'' \emph{Science Robotics}, vol.~2,
  no.~9, 2017. [Online]. Available:
  \url{https://robotics.sciencemag.org/content/2/9/eaan6357}
\BIBentrySTDinterwordspacing

\bibitem{hauser2018compliant}
S.~Hauser, M.~Mutlu, P.~Banzet, and A.~J. Ijspeert, ``Compliant universal
  grippers as adaptive feet in legged robots,'' \emph{Advanced Robotics},
  vol.~32, no.~15, pp. 825--836, 2018.

\bibitem{cianchetti2013stiff}
M.~Cianchetti, T.~Ranzani, G.~Gerboni, I.~De~Falco, C.~Laschi, and
  A.~Menciassi, ``Stiff-flop surgical manipulator: Mechanical design and
  experimental characterization of the single module,'' in \emph{2013 IEEE/RSJ
  international conference on intelligent robots and systems}.\hskip 1em plus
  0.5em minus 0.4em\relax IEEE, 2013, pp. 3576--3581.

\bibitem{wei2016novel}
Y.~Wei, Y.~Chen, T.~Ren, Q.~Chen, C.~Yan, Y.~Yang, and Y.~Li, ``A novel,
  variable stiffness robotic gripper based on integrated soft actuating and
  particle jamming,'' \emph{Soft Robotics}, vol.~3, no.~3, pp. 134--143, 2016.

\bibitem{zhu2019fully}
M.~Zhu, Y.~Mori, T.~Wakayama, A.~Wada, and S.~Kawamura, ``A fully
  multi-material three-dimensional printed soft gripper with variable stiffness
  for robust grasping,'' \emph{Soft robotics}, vol.~6, no.~4, pp. 507--519,
  2019.

\bibitem{stano2021}
G.~Stano and G.~Percoco, ``\BIBforeignlanguage{en}{Additive manufacturing aimed
  to soft robots fabrication: {{A}} review},''
  \emph{\BIBforeignlanguage{en}{Extreme Mechanics Letters}}, vol.~42, p.
  101079, Jan. 2021.

\bibitem{kapadia2012design}
J.~Kapadia and M.~Yim, ``Design and performance of nubbed fluidizing jamming
  grippers,'' in \emph{2012 IEEE International Conference on Robotics and
  Automation}.\hskip 1em plus 0.5em minus 0.4em\relax IEEE, 2012, pp.
  5301--5306.

\bibitem{li2017passive}
Y.~Li, Y.~Chen, Y.~Yang, and Y.~Wei, ``Passive particle jamming and its
  stiffening of soft robotic grippers,'' \emph{IEEE Transactions on Robotics},
  vol.~33, no.~2, pp. 446--455, 2017.

\bibitem{li2018distributed}
Y.~Li, Y.~Chen, and Y.~Li, ``Distributed design of passive particle jamming
  based soft grippers,'' in \emph{2018 IEEE International Conference on Soft
  Robotics (RoboSoft)}.\hskip 1em plus 0.5em minus 0.4em\relax IEEE, 2018, pp.
  547--552.

\bibitem{amend2016soft}
J.~Amend, N.~Cheng, S.~Fakhouri, and B.~Culley, ``Soft robotics
  commercialization: Jamming grippers from research to product,'' \emph{Soft
  robotics}, vol.~3, no.~4, pp. 213--222, 2016.

\bibitem{jiang2012learning}
Y.~Jiang, J.~R. Amend, H.~Lipson, and A.~Saxena, ``Learning hardware agnostic
  grasps for a universal jamming gripper,'' in \emph{2012 IEEE International
  Conference on Robotics and Automation}.\hskip 1em plus 0.5em minus
  0.4em\relax IEEE, 2012, pp. 2385--2391.

\bibitem{licht2017stronger}
S.~Licht, E.~Collins, M.~L. Mendes, and C.~Baxter, ``Stronger at depth: Jamming
  grippers as deep sea sampling tools,'' \emph{Soft robotics}, vol.~4, no.~4,
  pp. 305--316, 2017.

\bibitem{cheng2016prosthetic}
N.~Cheng, J.~Amend, T.~Farrell, D.~Latour, C.~Martinez, J.~Johansson,
  A.~McNicoll, M.~Wartenberg, S.~Naseef, W.~Hanson, \emph{et~al.}, ``Prosthetic
  jamming terminal device: A case study of untethered soft robotics,''
  \emph{Soft robotics}, vol.~3, no.~4, pp. 205--212, 2016.

\bibitem{athanassiadis2014particle}
A.~G. Athanassiadis, M.~Z. Miskin, P.~Kaplan, N.~Rodenberg, S.~H. Lee,
  J.~Merritt, E.~Brown, J.~Amend, H.~Lipson, and H.~M. Jaeger, ``Particle shape
  effects on the stress response of granular packings,'' \emph{Soft Matter},
  vol.~10, no.~1, pp. 48--59, 2014.

\bibitem{delaney2010packing}
G.~W. Delaney and P.~W. Cleary, ``The packing properties of superellipsoids,''
  \emph{EPL (Europhysics Letters)}, vol.~89, no.~3, p. 34002, 2010.

\bibitem{miskin2013adapting}
M.~Z. Miskin and H.~M. Jaeger, ``Adapting granular materials through artificial
  evolution,'' \emph{Nature materials}, vol.~12, no.~4, pp. 326--331, 2013.

\bibitem{jaeger2015celebrating}
H.~M. Jaeger, ``Celebrating soft matter’s 10th anniversary: Toward jamming by
  design,'' \emph{Soft matter}, vol.~11, no.~1, pp. 12--27, 2015.

\bibitem{RN406}
G.~W. Delaney, D.~Howard, and K.~De~Napoli, ``Utilising evolutionary algorithms
  to design granular materials for industrial applications,'' \emph{2019 18th
  IEEE International Conference On Machine Learning And Applications (ICMLA)},
  pp. 1897--1902, 2019.

\bibitem{jiang2014robotic}
A.~Jiang, T.~Ranzani, G.~Gerboni, L.~Lekstutyte, K.~Althoefer, P.~Dasgupta, and
  T.~Nanayakkara, ``Robotic granular jamming: Does the membrane matter?''
  \emph{Soft Robotics}, vol.~1, no.~3, pp. 192--201, 2014.

\bibitem{10.1145/3377929.3389951}
\BIBentryALTinterwordspacing
G.~W. Delaney and G.~Howard, ``Multi-objective exploration of a granular matter
  design space,'' in \emph{Proceedings of the 2020 Genetic and Evolutionary
  Computation Conference Companion}, ser. GECCO '20.\hskip 1em plus 0.5em minus
  0.4em\relax New York, NY, USA: Association for Computing Machinery, 2020, p.
  263–264. [Online]. Available: \url{https://doi.org/10.1145/3377929.3389951}
\BIBentrySTDinterwordspacing

\bibitem{RN414}
D.~Howard, A.~E. Eiben, D.~F. Kennedy, J.-B. Mouret, P.~Valencia, and
  D.~Winkler, ``Evolving embodied intelligence from materials to machines,''
  \emph{Nature Machine Intelligence}, vol.~1, no.~1, pp. 12--19, 2019.

\end{thebibliography}


\end{document}